\documentclass[runningheads]{llncs}

\usepackage{eccv}

\usepackage{eccvabbrv}

\usepackage{graphicx}
\usepackage{amssymb}
\usepackage{amsmath}
\usepackage{stmaryrd}
\usepackage{booktabs}
\usepackage{multirow}
\usepackage{multicol}
\usepackage{subcaption}

\usepackage[accsupp]{axessibility}  

\usepackage[pagebackref,breaklinks,colorlinks,citecolor=eccvblue]{hyperref}

\usepackage{orcidlink}

\newcommand{\exact}{\textsc{ExAct} }

\usepackage{amsmath,amsfonts,bm}

\def\eqref#1{equation~\ref{#1}}

\def\1{\bm{1}}

\def\va{{\bm{a}}}

\def\vs{{\bm{s}}}

\def\vz{{\bm{z}}}

\DeclareMathAlphabet{\mathsfit}{\encodingdefault}{\sfdefault}{m}{sl}
\SetMathAlphabet{\mathsfit}{bold}{\encodingdefault}{\sfdefault}{bx}{n}

\def\sD{{\mathbb{D}}}

\newcommand{\Ls}{\mathcal{L}}
\newcommand{\R}{\mathbb{R}}

\DeclareMathOperator*{\argmin}{arg\,min}

\begin{document}

\title{Towards Human Motion World Models \\ via Executable Behaviour Representations} 

\titlerunning{Executable Behaviour Representations}

\author{Rimvydas Rubavicius~\orcidlink{0000-0003-4270-7154} \and
Manisha Dubey~\orcidlink{0000-0002-2412-5137} \and
\\ N. Siddharth~\orcidlink{0000-0003-4911-7333} \and
Subramanian Ramamoorthy~\orcidlink{0000-0002-6300-5103} 
}

\authorrunning{R.~Rubavicius et al. 2026}

\institute{School of Informatics, University of Edinburgh \\
10 Crichton Street, Edinburgh EH8 9AB, UK \\
\email{rimvydas.rubavicius@ed.ac.uk}}

\maketitle

\begin{abstract}
Human motion world models should capture motion's intentionality by being executable: adaptable to different actions and capable of assessing motion quality. To achieve this, we introduce a domain-specific language \exact that represents human motions as underspecified programs that can be compiled to a reward model for zero-shot policy inference. By leveraging the compositional nature of \exact programs, we combine individual policies into executable behaviour representations. We evaluate the utility of the proposed approach by analysing human motion capture for the tasks of human action segmentation and human action anomaly detection. Our results suggest that the improvement in data efficiency and the capture of intuitive relationships between human actions are better than those of task-specific models.\footnote{Project website: \url{https://github.com/assistive-autonomy/exact}}
\end{abstract}

\begin{figure*}[h!]
    \centering
    \includegraphics[width=0.7\linewidth]{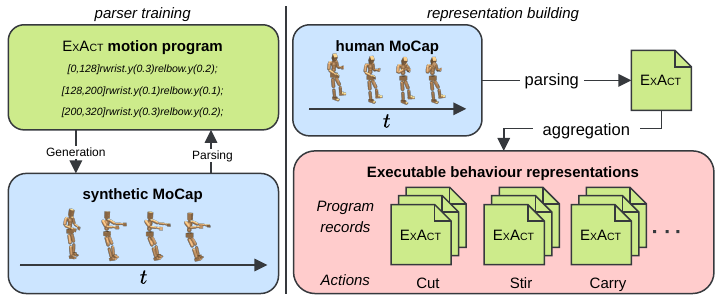}
    \caption{Overview of executable behaviour representations. The parser is trained using synthetically generated data. It is used in representation building to parse and aggregate programs from human motion capture based on action labels.}
    \label{fig:overview}
\end{figure*}

\section{Introduction}
\label{sec:introduction}

Consider a cook making a sauce. Their stirring exhibits a circular motion that not only whisks but also reflects adaptation to the pot's dimensions and heat intensity. A pause may signal a new action, such as heat adjustment. While the action is categorised as ``stir'', this label does not convey the action's mechanics. Human motions are intentional and structured: they are carried out in the environment to achieve a goal following a motion sequence~\cite{BAKER2009329}. Contemporary approaches to understanding humans can recognise actions and anticipate changes in them from observations: they recognise \textit{what} actions are performed as action labels~\cite{DBLP:journals/firai/VrigkasNK15}, which in turn have been used to estimate plans and goals~\cite{DBLP:series/synthesis/2021Mirsky}.
Nevertheless, such representation is limited, as it does not assess \textit{how} action is performed; yet such knowledge is necessary to assess the quality of the action.

To this end, we study learning a component of a human motion world model---executable behaviour representations (EBRs)---from human motion capture (MoCap) data (see Fig.~\ref {fig:overview}). We introduce an~\exact domain-specific language that represents human motion as programs. It uses keypoints as primitives to describe human motion in an underspecified manner: they do not explicitly prescribe behaviour for each keypoint or describe the action's geometry. This unique feature allows programs to be interpreted as reward models~\cite{DBLP:journals/natmi/DavidsonTTGL25} for zero-shot policy inference~\cite{DBLP:journals/corr/abs-2101-07123,DBLP:conf/iclr/TirinzoniTFGKXL25}. Using \exact, we generate synthetic program-motion pairs and train a parser from MoCap to programs. Using programs, we describe a procedure for representation building that aggregates programs by action labels into EBR. We evaluate our approach on three datasets from two benchmarks: HumanAct12~\cite{guo2020action2motion}, ESK Activities, and ESK Verbs~\cite{DBLP:journals/corr/abs-2506-01608}, using action labels but without access to the ground-truth programs. We evaluate EBR on the tasks of human action segmentation and human action anomaly detection. Our results suggest that EBR achieve greater data efficiency and compositional generalisation than task-specific models.

\section{Related Work}
\label{sec:related_work}

Motion programs~\cite{DBLP:conf/cvpr/KulalMA021,DBLP:conf/cvpr/KulalMA022} have been used for recognition and generation based on a three-level hierarchy of human poses, motion concepts, and activity programs, with learning performed to estimate motion concepts from human poses and to synthesise programs from motion concepts. Our approach differs from such programs in structure and paradigm. For structure, our programs are \textit{concept-free} and are directly constructed from keypoints. We achieve this by basing \exact predicates on keypoints of human poses that constitute human motion, while omitting other keypoints due to underspecification. This allows us to consider a wider range of motions, not constrained by learned or engineered motion concepts. In particular, even though \exact programs are based on the SMPL~\cite{SMPL:2015}, \exact can, in principle, be changed to other kinematic models. For paradigm, \exact captures the intentionality of human motion, being functional and declarative, while prior programs are imperative and procedural.

World Models~\cite{3327144.3327171,DBLP:journals/corr/abs-2306-02572,DBLP:journals/nature/HafnerPBL25} are learned environment representations: they abstract the complexity of the observations but still capture the environment's dynamics. World models have three functional components: a renderer for predicting future observations, a simulator for predicting future states, and a planner for predicting actions~\cite{li2026taxonomy}. We define a human motion world model (HuMoWM) as a world model of the dynamics of the human body. Prior work on human motion modelling focuses on human motion generation with~\cite{DBLP:conf/cvpr/GuoZZ0JL022,zhang2023generating,DBLP:conf/cvpr/GuoMJW024} or without~\cite{rempe2021humor,Luo2023PerpetualHC,luo2024universal} conditioning of language prompts, which in turn contributes to developing the rendering and simulation component for HuMoWMs. In this work, we focus on developing the planning component for building EBR. Executability~\cite{Fisher2007ExecutableCB,Clarke2020ExecutableCM} is used to describe models that capture the internal structure of a system as its surrogate, mimicking the system's underlying computational model rather than focusing on perceptual realism. In this paper, we build EBR to programmatically control SPML rather than raw observations.

\section{Executable Behaviour Representations (EBR)}

We consider the SMPL humanoid~\cite{SMPL:2015} with 23 joints. Its state~\(\vs\in\R^{358}\) includes proprioception, which positions~\(\vs^{\mathrm{pos}}\in\R^{69}\) equivalent to MoCap and velocities~\(\vs^{\mathrm{vel}}\in\R^{69}\) in 3D directions with respect to the humanoid's frame while control actions~\(\va\in[-1,1]^{69}\) describe the forces for each joint in 3D directions. Motion \(\vs_{0:T}=\vs_0,\dots,\vs_{T}\) is a sequence of states in the environment as generated by action sequences~\(\va_{1:T-1}=\va_1,\dots,\va_{T-1}\). Successor representation~\cite{succesor_dayan} is a distribution over future states following a policy, representing the coupling between the behaviour and the environment. It can be factored into forward-backwards representations~\cite{DBLP:journals/corr/abs-2101-07123} in which states, actions, and rewards are in the same latent space, allowing zero-shot policy inference: given a buffer \(\sD\) of states-reward pairs~\((\vs,r)\), the action-value function is given in a closed form:
\begin{equation}
        Q(\vs,\va) = F(\vs, \va)^\top \vz; \qquad \vz = \frac{1}{|\sD|} \sum_{\vs,r\in\sD} rB(\vs)
\end{equation}
where \(F\) and \(B\) are forward-backwards representations, which, in the scope of this paper, are neural networks, optimised using human MoCap and a buffer of 500K state-reward pairs~\cite{DBLP:conf/iclr/TirinzoniTFGKXL25}. They encode a prior for human-like motions and allow underspecified rewards: rewarding only for reaching particular state values while other state values follow behaviour in the successor representation.

We design \exact using the SMPL joints as predicates. \exact syntax is expressed using a BNF grammar for the recursive structure of programs~\(\phi\), with \(\langle\cdot\rangle\) denoting non-terminals in \exact grammar:
\begin{align}
\begin{split}
\label{eq:syntax}
 \langle \text{program} \rangle &::= \langle \text{motion} \rangle | \langle \text{motion} \rangle ;\langle \text{motion} \rangle \\
    \langle \text{motion} \rangle &::= \texttt{[} t_1,t_2
    \texttt{]} \, \langle \text{sensors} \rangle \\
    \langle \text{sensors}\rangle &::= \langle \text{sensor} \rangle | \langle \text{sensor} \rangle \langle \text{sensor} \rangle \\
    \langle \text{sensor} \rangle &::= \mathtt{s^{pos}} \texttt{(} x\texttt{)} 
\end{split}
\end{align}

where \(t_1<t_2<T\in\) are start and end timesteps and \(x\in[-1,1]\) is the target position. Programs have one or more motions, each with a start and end timestep. For example, \([5]\mathtt{LArm.x}(0.3)\) is a program that represents a single motion for 5 timesteps, while \([0,10]\mathtt{LArm.x}(0.3);[10,20]\mathtt{RArm.y}(0.1)\) represents two motions, each of 10 timesteps. For each motion, there exists one or more position sensors~\(\mathtt{s}^{\mathrm{pos}}\) like \(\mathtt{RArm.y}\) that represent changing the yaw of the right arm and are optimised to reach the target position. \exact programs are compiled to a set of \(\vz_t\) for timesteps for target motions. To this end, we utilise a buffer~\(\sD\) that encodes partial reward assignments upon reaching the target. We use \exact grammar to recursively define \exact semantics:
\begin{align}
\begin{split}
\llbracket\langle\text{program}\rangle\rrbracket &= \bigcup_{\langle\text{motion}\rangle\in \langle\text{program}\rangle}\llbracket\langle\text{motion}\rangle\rrbracket \\ 
    \llbracket\langle \text{motion}\rangle\rrbracket &= \vz_{t_1:t_2} = \frac{1}{|\sD|}\sum_{s,r\in\sD} \llbracket\langle\text{sensors}\rangle\rrbracket B(s) \\  
\llbracket\langle\text{sensors}\rangle\rrbracket &= \prod_{\langle\text{sensor}\rangle\in \langle\text{sensors}\rangle}\llbracket\langle\text{sensor}\rangle\rrbracket \\ 
    \llbracket\langle\text{sensor}\rangle\rrbracket &= \sigma((s^{\mathrm{pos}} - x)^2) 
\end{split}
\end{align}
where \(\vz_{t_1:t_2}\) applies for \(t_1:t_2\) timesteps, \(\sigma\) is a sigmoid,  and \(s^{\mathrm{pos}}\) is the value of joint position \(\mathtt{s}^{\mathrm{pos}}\) for the state~\(s\) in \(\sD\). The interpretation of each program~\(\phi\) is a set of the \(\vz\) that be used to compute action-value function:
\begin{equation}
\label{eq:q_phi}
    Q_\phi(\vs_{t},\va_{t}) = F(\vs_t,\va_t)^\top\vz_t^{\phi}; \quad
    \vz^\phi_t = \begin{cases}
    \vz_{t_1:t_2} &\text{for }t\in[t_1,t_2]\text{ and }\vz_{t_1:t_2} \in\llbracket\phi\rrbracket \\
    0 &\text{otherwise}
    \end{cases}
\end{equation}

The parser follows an encoder-decoder architecture. It consists of a spatio-temporal graph convolutional network (ST-GCN) encoder (good for exploiting the SMPL skeleton structure~\cite{10.5555/3504035.3504947}) that embeds MoCap~\(\vs^{\rm pos}_{0:T}\) into the latent space of motion tokens. The generated motion tokens are used as a prefix~\cite{DBLP:conf/nips/LiuLWL23a} for an autoregressive decoder, which, in our case, is Qwen2.5-Coder-3B~\cite{DBLP:journals/corr/abs-2409-12186}. In the encoder, all parameters are trainable, while for the decoder, only a small portion is trainable using LoRA~\cite{DBLP:conf/iclr/HuSWALWWC22}. To learn the parser, we use the \exact grammar (Eq.~\ref{eq:syntax}) and forward-backwards representations~\cite{DBLP:conf/iclr/TirinzoniTFGKXL25} to produce a dataset of program-motion pairs. Note that we use only the positional component of the state~\(\vs^{\rm pos}\subset\vs\) for training, since it is equivalent to MoCap. The parser is trained by minimizing weighted loss~\(\Ls = \Ls_{\rm LM} + \lambda\Ls_{\rm InfoNCE}\) with \(\Ls_{\rm LM}\) measuring the difference between predicted and target programs and \(\Ls_{\rm InfoNCE}\)~\cite{DBLP:journals/jmlr/GutmannH10,DBLP:journals/corr/abs-1807-03748} used to align the motion tokens produced by the encoder to the latent space of the language tokens. To ensure program validity, we use grammar-guided generation~\cite{DBLP:journals/tmlr/UgareSKM025}.

\exact programs are temporally and functionally compositional: motion can be sequentially added to generate motion sequences, and multiple programs can, in disjunction, influence the reward. This allows us to aggregate the set of programs \(\Phi\) for each action. To compose programs into an EBR, we perform the arithmetic logical disjunction of \(\vz\)'s of programs:

\begin{equation}
    Q_{\Phi}(\vs_{t},\va_{t}) = F(\vs_t,\va_t)^\top\vz^\Phi_t;  \qquad \vz^\Phi_t = 1 - \sum_{\phi\in\Phi} (1-\vz^\phi_t)
\end{equation}

Note that the number of programs~\(|\Phi|<100\) is bounded, with the selection maximising diversity, measured by program length and predicate variety.

\section{Experiments}
\label{sec:experiments}

We use NVIDIA A100 80GB for up to 300 compute hours, including both parser training and experiments using two benchmarks: 1) HumanAct12~\cite{guo2020action2motion} of 90K poses annotated with 12 action labels (\eg ``walk'', ``run'') from 12 subjects (7 for training, 1 for validation, and 6 for testing); 2) ESK~\cite{DBLP:journals/corr/abs-2506-01608} of 16 subjects cooking 4 different recipes with a total of 51 sessions recorded (33 for training, 2 for validation, 16 for testing). We consider two datasets that annotate the same sessions but with different granularity: ESK Activities of coarse-grained actions (\eg ``cooking'', ''cleaning'') and ESK Verbs of fine-grained actions (\eg ``add'', ``cut''). For parser training, we generate a set of 50K programs with varying numbers of predicates, program length and time horizon. The programs are used to generate synthetic MoCap using forward-backwards representations~\cite{DBLP:conf/iclr/TirinzoniTFGKXL25}. We also generate 3K program-motion pairs for Bayesian optimisation in sweeps for hyperparameter tuning (training for 5 epochs for 10 trials) for parser training (learning rate, weight decay, warm-up steps, gradient accumulation steps, batch size)ST-GCN encoder  (number of layers, their dimensionality, and the motion token size), and LoRA (rank, alpha, dropout) set as part of hyperparameter sweeps. After initial training runs, in which we evaluate the scales and importance of two losses, we set \(\lambda=0.15\) before the hyperparameter sweeps.

To evaluate EBR, we use the human action segmentation task, in which, given an SMPL motion sequence, we assign an action label to each timestep. We use EBR for data augmentation by generating synthetic MoCap data from initial human poses. We use the DLC2Action~\cite{Kozlova2025.09.27.678941} to evaluate a variety of action segmentation models. Following Bonnetto et al.~\cite{DBLP:journals/corr/abs-2506-01608} we consider four pose-based action segmentation models: MS-TCN3~\cite{DBLP:journals/pami/LiFLCG23}, EDTCN~\cite{DBLP:conf/cvpr/LeaFVRH17}, C2F-TCN~\cite{singhania2022iterative}, and C2F-TRF (C2F-TCN with convolutions replaced with attention). We use DLC2Action to extract features of speed, acceleration, speed direction, joint angles, joint acceleration, intra-coordinate distance, distances to centroids, and centroid positions, as is done for ESK pose-based action segmentation baselines (see Appendix of \cite{DBLP:journals/corr/abs-2506-01608}). The key difference between our setup and that reported in Bonnetto et al.~\cite{DBLP:journals/corr/abs-2506-01608} is that we use the SMPL rather than the simpler 17-keypoint model to match \exact grammar. Note that both HumanAct12 and ESK natively record human pose information in SMPL. We consider segmentation models trained on three types of training data: 1) original: original dataset training data; 2) +perturbed: original data and data produced with default data augmentations in the DLC2Action (mirroring, shifting, and adding noise); 3) +synthetic: original data and 1K exemplars generated using EBR. When training the segmentation models, we perform hyperparameter sweeps (10 epochs across 20 trials) and select the model based on the validation F1. With the best hyperparameters, we train each segmentation model for 3 runs over 50 epochs and record segmental F1 to measure the model's effectiveness in determining action boundaries and PR-AUC to measure frame-level classification.

Table~\ref{tab:action_segmentation_results} shows segmentation experiment results. We observe that +synthetic often leads to gains without a performance decrease, unlike what was observed for +perturbed. We attribute the +perturbed performance decrease to a breakdown in temporal dependencies in motion. Nevertheless, +perturbed is useful for performance gains on the PR-AUC, particularly for ESK Activities. This can be partially explained by the lower number of action labels, as performance degrades on datasets with more labels. When analysing programs generated in dataset parsing, we observe that our parser, although trained on a diverse set of programs with multiple motions, tends to produce simpler programs (containing one or two motion sequences). This, in turn, explains the lack of significant gains from using EBR for the ESK Activities, as data augmentation cannot capture the complexity of temporally extended actions.

\begin{table}[t]
\centering
\caption{Human action segmentation results averaged over 3 runs, reporting Segmental F1 and PR-AUC. Bold indicates the best statistically significant training-data variant.}
\label{tab:action_segmentation_results}
\setlength{\tabcolsep}{4pt}
\resizebox{\textwidth}{!}{%
\begin{tabular}{@{}cccccccc@{}}
\toprule
\multirow{2}{*}{\textbf{Dataset}} & \multirow{2}{*}{\textbf{\begin{tabular}[c]{@{}c@{}}Segmentation\\ Models\end{tabular}}} & \multicolumn{3}{c}{\textbf{Segmental F1}} & \multicolumn{3}{c}{\textbf{PR-AUC}} \\
\cmidrule(lr){3-5} \cmidrule(lr){6-8}
& & original & +perturbed & +synthetic & original & +perturbed & +synthetic \\ \midrule
\multirow{4}{*}{HumanAct12}
& MS-TCN3 & 27.42\(\pm0.61\) & 24.04\(\pm2.40\) & 23.04\(\pm1.51\) & 14.06\(\pm0.67\) & 17.37\(\pm0.12\) & \textbf{19.65\(\pm0.06\)} \\
& C2F-TCN & 28.16\(\pm1.80\) & 27.90\(\pm1.28\) & \textbf{32.43\(\pm0.76\)} & 42.71\(\pm0.12\) & 43.43\(\pm0.63\) & \textbf{54.86\(\pm3.15\)} \\
& EDTCN & 22.78\(\pm1.22\) & 18.24\(\pm0.76\) & \textbf{29.34\(\pm1.03\)} & 14.67\(\pm0.22\) & 14.20\(\pm0.11\) & 13.93\(\pm0.37\) \\
& C2F-TRF & 37.48\(\pm3.01\) & 34.00\(\pm0.92\) & 35.37\(\pm0.77\) & 42.38\(\pm0.85\) & \textbf{47.44\(\pm0.09\)} & 15.35\(\pm0.46\) \\ \midrule
\multirow{4}{*}{\begin{tabular}[c]{@{}c@{}}ESK\\ Activities\end{tabular}}
& MS-TCN3 & \textbf{18.27\(\pm2.72\)} & 05.71\(\pm0.78\) & 13.49\(\pm2.22\) & 31.39\(\pm2.16\) & 31.59\(\pm2.31\) & 19.14\(\pm0.65\) \\
& C2F-TCN & 24.81\(\pm1.81\) & 19.96\(\pm2.17\) & 24.66\(\pm0.49\) & 21.69\(\pm1.71\) & \textbf{51.45\(\pm1.90\)} & 18.89\(\pm0.18\) \\
& EDTCN & 10.77\(\pm0.26\) & \textbf{15.32\(\pm0.16\)} & 11.18\(\pm0.30\) & 14.35\(\pm0.08\) & \textbf{21.90\(\pm0.39\)} & 14.18\(\pm0.29\) \\
& C2F-TRF & 26.15\(\pm0.37\) & 23.02\(\pm1.28\) & 24.31\(\pm1.01\) & 21.18\(\pm0.24\) & \textbf{30.49\(\pm2.87\)} & 21.60\(\pm0.61\) \\ \midrule
\multirow{4}{*}{\begin{tabular}[c]{@{}c@{}}ESK\\ Verbs\end{tabular}}
& MS-TCN3 & \textbf{07.79\(\pm0.13\)} & 01.30\(\pm0.22\) & 06.49\(\pm0.61\) & 11.13\(\pm0.12\) & 09.93\(\pm0.70\) & 11.37\(\pm0.22\) \\
& C2F-TCN & 06.55\(\pm0.25\) & 05.88\(\pm0.19\) & 05.84\(\pm0.33\) & 11.50\(\pm0.06\) & 11.22\(\pm0.28\) & 11.47\(\pm0.23\) \\
& EDTCN & 06.89\(\pm0.32\) & 01.53\(\pm0.72\) & \textbf{07.46\(\pm0.02\)} & \textbf{08.38\(\pm0.08\)} & 03.40\(\pm0.39\) & \textbf{08.37\(\pm0.13\)} \\
& C2F-TRF & 05.86\(\pm0.36\) & 05.74\(\pm0.12\) & \textbf{07.24\(\pm0.27\)} & 18.30\(\pm0.24\) & 12.97\(\pm0.15\) & \textbf{20.25\(\pm0.17\)} \\ \bottomrule
\end{tabular}%
}
\end{table}

\begin{figure*}[h!]
    \centering
\includegraphics[width=0.7\textwidth]{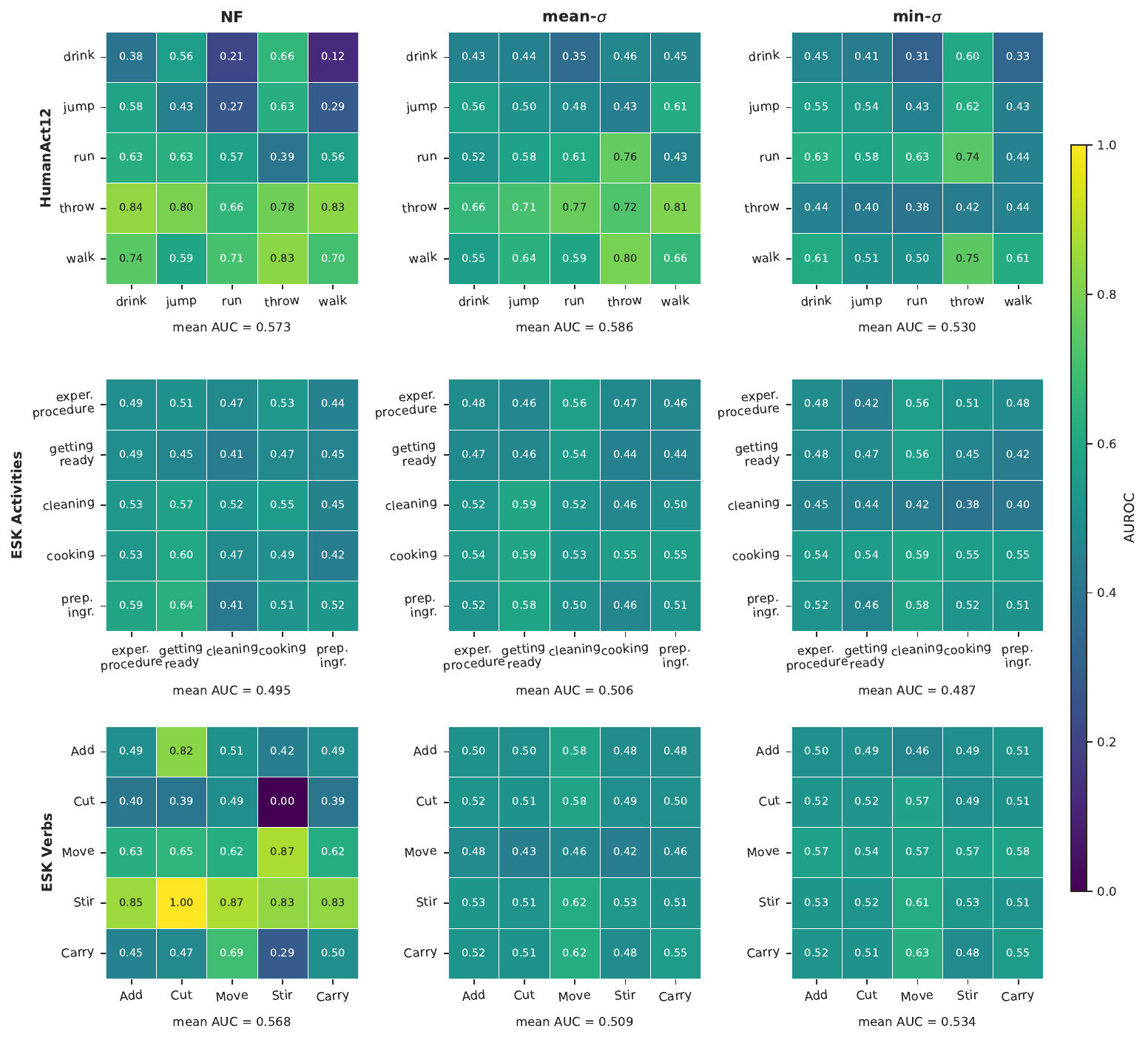}
    \caption{AUROC matrices for the 5 actions with target action (learning assessment model) on the y-axis and query action (evaluating assessment model) on the x-axis.}
    \label{fig:human_action_annomaly_detection}
\end{figure*}

EBR allows action assessment. We test this with a human action anomaly detection task, in which EBR estimates how similar MoCap is for different actions. We use the training data from each dataset to learn an assessment model for the target action like ``cut'', and use it to measure its similarity with the queries like \eg``add''. From each dataset, we use 5 actions. For the similarity measure using EBR, we use the unordered tree edit distance~\(d_{\mathrm{edist}}\)~\cite{DBLP:journals/siamcomp/ZhangS89} between the programs in EBR~\(\phi\in\Phi\) and the query action program~\(\psi\) as an activation for the sigmoid: \(\sigma(-d_{\mathrm{edist}}(\phi,\psi))\). We weight the edit costs, with the highest penalties for using the wrong side (left instead of right), wrong joints, and wrong directions, respectively. In \(\mathrm{edist}\), we also ignore the temporal scale. We consider three assessment models: 1) normalising-flow (NF): a density estimator for pose-based human action anomaly detection~\cite{DBLP:conf/iccv/HirschornA23} which uses an ST-GCN encoder to process MoCap and fit the Gaussian mixture model that allows estimating the likelihood for each query; 2) mean-sigmoid (mean-\(\sigma\)): a density estimator based on EBR that measures the mean response of the sigmoid with activation given as~\(\frac{1}{|\phi|}\sum_{\phi\in\Phi}\sigma(-d_{\mathrm{edist}}(\phi,\psi))\); 3) min-sigmoid (min-\(\sigma\)): a density estimator based on EBR that measures the minimum response of the sigmoid with activation given as~\(\argmin_{\phi\in\Phi}\sigma(-d_{\mathrm{edist}}(\phi,\psi))\). For evaluation, we use the mean AUC and the AUROC matrix, with the former capturing in-class and out-of-class differences to examine how a model trained on a single action relates to query actions.

Fig.~\ref{fig:human_action_annomaly_detection} shows the AUROC matrices. For the ESK Activity, we observe that none of the assessment models captures significant differences between actions, as they do not account for the full complexity of the action (NF is limited by the context window, and edit-distance-based models are limited by simple program synthesis). For HumanAct12, we observe that mean-\(\sigma\) yields the highest mean AUC. When analysing their AUROC, we see that they encode similarities between actions; for example, for ``walk''-``run'', we observe minimal differences between the programs (which is expected as the joints used are the same for motion generation), while NF ends up discriminating between the two actions, not capturing the semantic similarity between actions. This is expected for programs, as they are underspecified and do not account for the spurious cues for action differentiation and focus on joints used in motion generation, which are the same for ``walk'' and ``run''. Finally, for ESK Verbs, we observe a conservative separation profile similar to ``walk'' and ``run'' in HumanAct12, signalling high similarity between activities, captured by EBR. Finally, there are minimal differences in AUROC for mean-\(\sigma\) and min-\(\sigma\), indicating that there is low diversity between the target and query action programs.

\section{Conclusions}
\label{sec:conclusions}

We presented a method for learning EBR which can be used as a planner for HuMoWM. For future work, we consider joint learning other components for HuMoWM, in particular a renderer for handling multimodal observations.

\section*{Acknowledgements}

This work was supported by a UKRI Turing AI World Leading Researcher Fellowship on AI for Person-Centred and Teachable Autonomy (grant EP/Z534833/1), the Edinburgh International Data Facility (EIDF) and the Data-Driven Innovation Program at the University of Edinburgh. Access to EIDF was facilitated through the University of Edinburgh’s Generative AI Laboratory GAIL Fellow scheme.

%
%
\bibliographystyle{splncs04}
\bibliography{references}
\end{document}